\documentclass{article}

\usepackage{PRIMEarxiv}

\usepackage[utf8]{inputenc} 
\usepackage[T1]{fontenc}    
\usepackage{hyperref}       
\usepackage{url}            
\usepackage{booktabs}       
\usepackage{amsfonts}       
\usepackage{nicefrac}       
\usepackage{microtype}      
\usepackage{graphicx}
\usepackage{natbib}         
\usepackage{doi}
\usepackage{multirow}
\usepackage{multicol}
\usepackage{amsmath}

\title{Hybrid Fusion: One-Minute Efficient Training for Zero-Shot Cross-Domain Image Fusion}
\author{ 
  Ran Zhang \\
  Hefei University of Technology \\
  Hefei, China \\
  \And
  Xuanhua He \\
  University of Science and Technology of China \\
  Hefei, China \\
  \And
  Liu Liu \\
  Hefei University of Technology \\
  Hefei, China \\
}

\begin{document}






\maketitle

\begin{abstract}
Image fusion seeks to integrate complementary information from multiple sources into a single, superior image. While traditional methods are fast, they lack adaptability and performance. Conversely, deep learning approaches achieve state-of-the-art (SOTA) results but suffer from critical inefficiencies: their reliance on slow, resource-intensive, patch-based training introduces a significant gap with full-resolution inference. We propose a novel hybrid framework that resolves this trade-off. Our method utilizes a learnable U-Net to generate a dynamic guidance map that directs a classic, fixed Laplacian pyramid fusion kernel. This decoupling of policy learning from pixel synthesis enables remarkably efficient full-resolution training, eliminating the train-inference gap. Consequently, our model achieves SOTA-comparable performance in about \textbf{one minute} on a RTX 4090 or two minutes on a consumer laptop GPU from scratch without any external model and demonstrates powerful zero-shot generalization across diverse tasks, from infrared-visible to medical imaging. By design, the fused output is linearly constructed solely from source information, ensuring high faithfulness for critical applications. The codes are available at \url{https://github.com/Zirconium233/HybridFusion}.
\end{abstract}

\section{Introduction}
\label{sec:intro}

\begin{figure}[t]
    \centering
    \includegraphics[width=0.95\columnwidth]{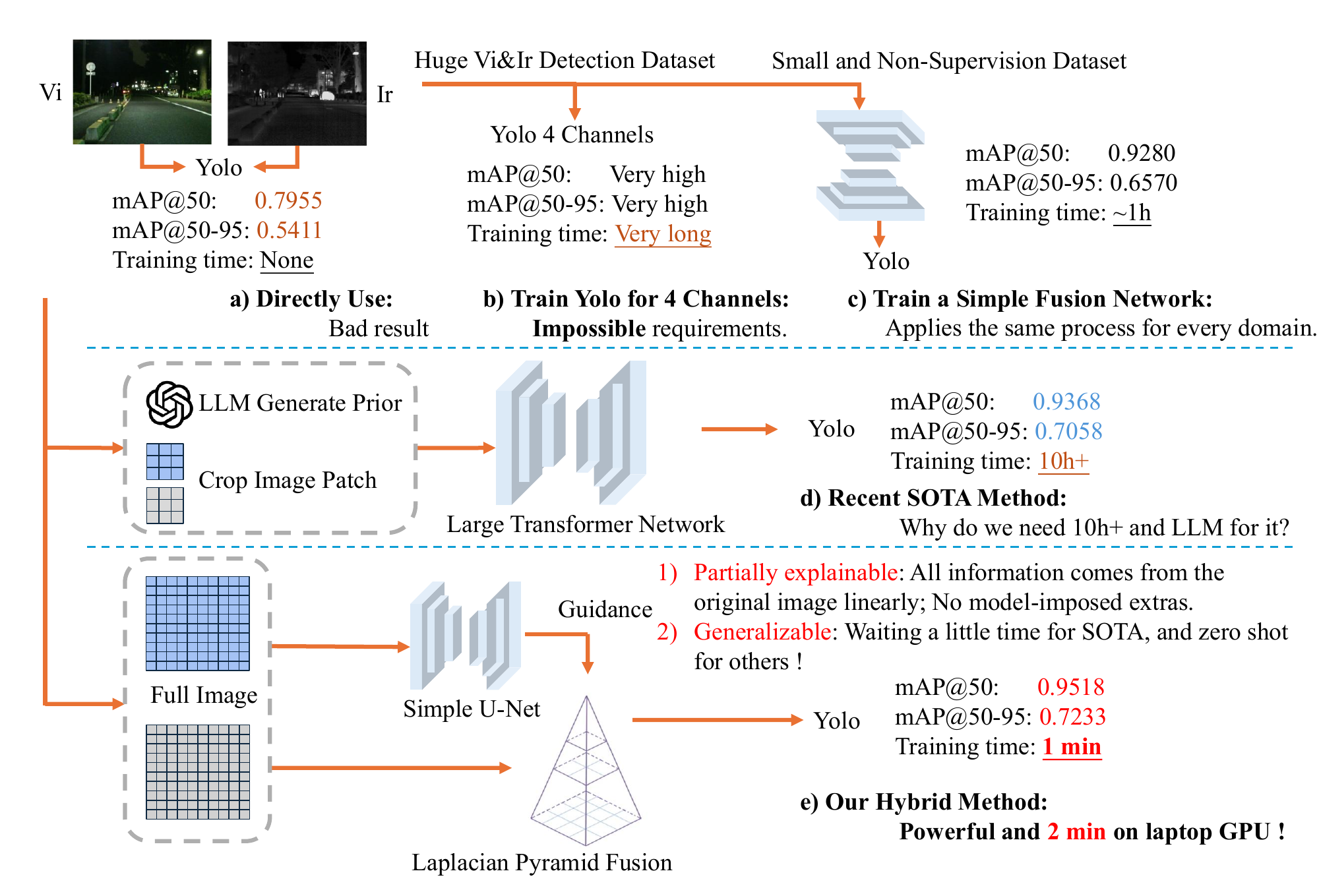}
    \caption{A conceptual comparison of image fusion paradigms. }
    \label{fig:fusion_paradigms}
    
\end{figure}

Image fusion aims to integrate complementary information from multiple source images into a single, superior composite that is more suitable for human perception and computer vision tasks~\cite{Albarqaan2024Image, MI}. Its applications are widespread, ranging from multi-spectral remote sensing to improving diagnostics in medical imaging by combining modalities like MRI and PET~\cite{xu2021emfusion, zhao2023cddfuse}.

Traditional fusion techniques were long dominated by multi-scale decomposition methods, such as those using Laplacian or wavelet pyramids, which merge image components with handcrafted rules~\cite{MI, laplacianPyramidFusionNetwork}. While computationally simple, these rules lack adaptability and often introduce visual artifacts~\cite{TANG2022SuperFusion}. The advent of deep learning offered a path to learn fusion strategies directly from data, starting from early convolutional frameworks like IFCNN~\cite{zhang2020IFCNN} and GAN-based models~\cite{ma2019fusionGAN} to unified architectures like U2Fusion~\cite{xu2020u2fusion}. More recently, hybrid designs such as HG-LPFN~\cite{laplacianPyramidFusionNetwork} sought to blend classic structures with neural networks. However, these methods often remain overly complex, treating classic techniques as mere feature engineering steps for networks that are still responsible for intensive pixel-level synthesis.

The primary limitation of these increasingly complex state-of-the-art (SOTA) methods~\cite{zhao2023cddfuse, zhao2023DDFM, ma2022swinfusion} is their reliance on patch-based training to manage immense memory consumption, creating a significant ``training-inference gap'' when applied to full-resolution images. Furthermore, recent works like Text-IF~\cite{yi2024text} and DTPF~\cite{dtpf} incorporate external priors from large-scale models to boost scores. While innovative, this synthesis-heavy approach can introduce hallucinations information not present in the original sources, which is a critical flaw for medical imaging where data faithfulness is paramount. Above all, the training efficiency of these models is rarely addressed; they often require hours or days to train~\cite{yi2024text}.

We believe that image fusion should not require massive computational overhead; indeed, many SOTA models now possess parameter counts that rival the size of their training datasets~\cite{dtpf, yi2024text}. Crucially, we argue that a fusion model does not need to learn image synthesis from scratch like a generative model~\cite{zhao2023DDFM}. Forcing a network to understand textures or natural image statistics on limited datasets often leads to hallucinations and restricted generalization. This inspired us to propose a ``linear'' hybrid paradigm where the model serves only as a guide for a traditional fusion algorithm. By decoupling policy learning from pixel synthesis, the model learns how to allocate source information rather than how to synthesize pixels. Unlike previous hybrid methods~\cite{laplacianPyramidFusionNetwork} that use pyramids as internal components, our approach uses the traditional kernel as a safety fallback, ensuring that even under poor training conditions, the output remains a valid fusion without noise or artifacts. This design allows the framework to reach SOTA-level performance with lightning-fast convergence and remarkable zero-shot generalization.

Our contributions are summarized as follows:
\begin{enumerate}

    \item We introduce a novel hybrid architecture where a U-Net only predicts a control map while a traditional algorithm executes the fusion. This eliminates the training-inference gap by enabling efficient end-to-end training on full-resolution images.

    \item Our method achieves unprecedented training efficiency, reaching competitive performance in approximately two minutes on a consumer-grade GPU (see Table~\ref{tab:performance_benchmarks}), a task for which other methods can take hours.

    \item We demonstrate strong zero-shot generalization, where a model trained only on natural scenes (MSRS dataset) achieves excellent performance on unseen medical imaging tasks (Table~\ref{tab:med_comparison} and \ref{tab:zeroshot_generalization}), while ensuring faithfulness to the source images.

\end{enumerate}

\section{Related Work}
\label{sec:related}

\paragraph{Traditional and Hybrid Methods.}
Multi-scale decomposition, such as Laplacian~\cite{laplacianCNN} and wavelet transforms, provides a classic framework for merging images via handcrafted rules~\cite{MI}. While interpretable, these fixed rules lack the adaptability required for diverse scenes. Early hybrid designs like HG-LPFN~\cite{laplacianPyramidFusionNetwork} integrated Laplacian pyramids with learned components; however, they typically treat the pyramid as an internal layer for non-linear pixel synthesis. This results in excessive training requirements (e.g., 10,000 epochs reported) and inherits the complexity of purely deep methods. In contrast, our framework treats the Laplacian pyramid as the fundamental architectural framework and delegates the entire synthesis process to the fixed kernel, simplifying the network's task to pure weight-map guidance.

\paragraph{Deep Learning-based Fusion.}
Deep learning has advanced fusion through automated strategy learning, progressing from GANs~\cite{ma2019fusionGAN} to Transformer-based backbones~\cite{ma2022swinfusion, zhao2023cddfuse, Zamir2021Restormer}. Recent SOTA methods further incorporate external priors~\cite{yi2024text}, diffusion models~\cite{zhao2023DDFM}, or large-model distillation~\cite{dtpf}. Despite their performance, these methods incur heavy computational costs, often requiring hours or days for training~\cite{yi2024text}. Moreover, to manage VRAM, they typically rely on patch-based training, creating a significant ``training-inference gap'' at full resolutions. By learning a control map rather than raw pixels, our method enables rapid, full-resolution training using a lightweight U-Net, effectively sidestepping the training bottlenecks of heavy reconstruction-based backbones.

\section{Method}
\label{sec:method}

\begin{figure*}[t]
    \centering
    \includegraphics[width=\textwidth]{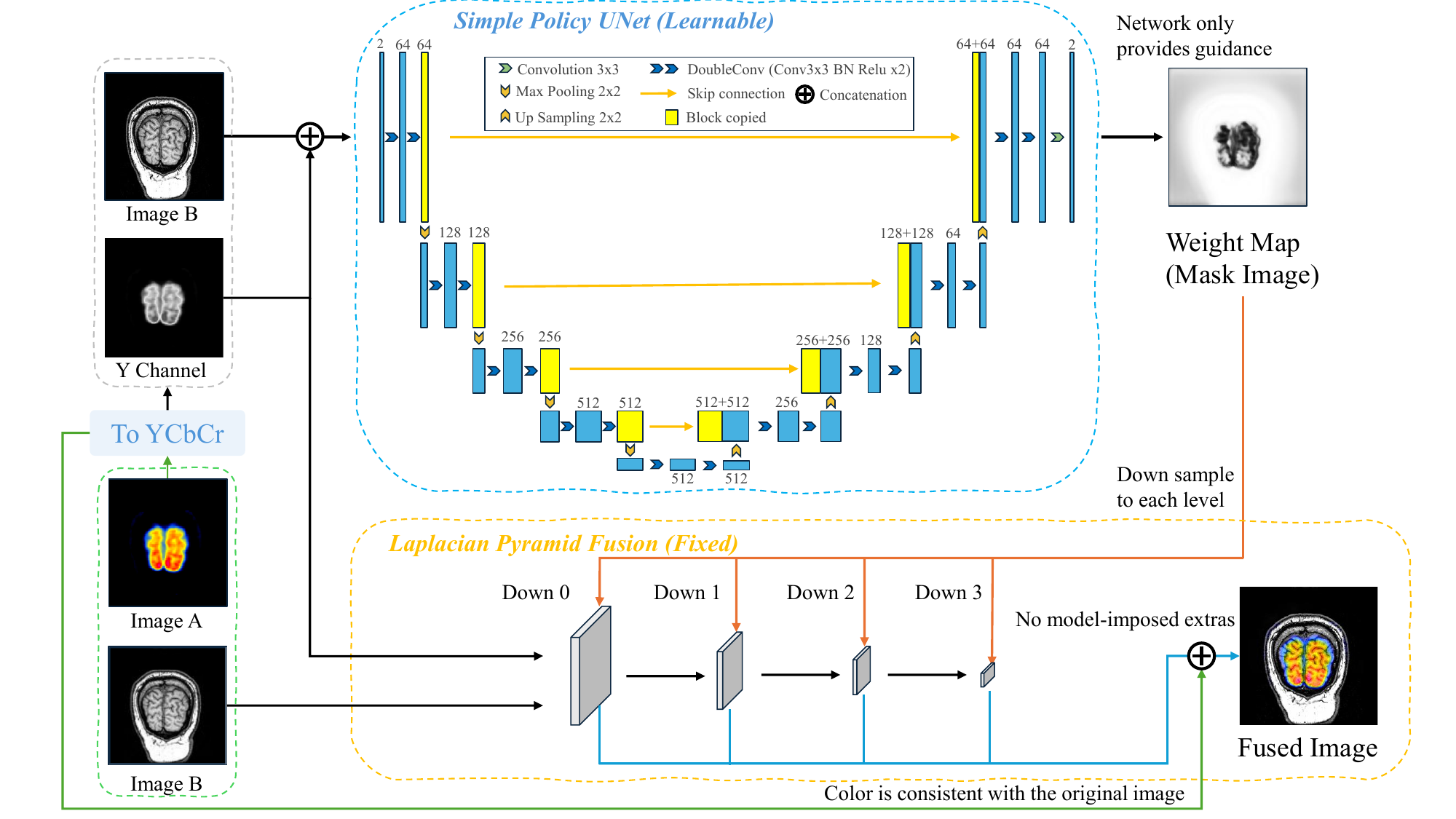}
    \caption{The architecture of our hybrid fusion model. A learnable U-Net (top) takes the concatenated visible luminance (Y) and infrared (Ir) channels to produce a guidance weight map. This map then directs a fixed, non-learnable Laplacian Pyramid fusion kernel (bottom) that operates on the multi-scale decompositions of the source images. The original chrominance (CbCr) is preserved and reapplied to the fused luminance, ensuring color faithfulness in the final output.}
    \label{fig:model_architecture}
    
\end{figure*}

Our proposed hybrid fusion framework is designed to be efficient, generalizable, and faithful to the source images. As illustrated in Figure~\ref{fig:model_architecture}, the architecture strategically decouples the fusion process into a learnable guidance generation stage and a fixed, traditional fusion stage. This separation is the key to its efficiency and effectiveness.

\subsection{Overall Pipeline}
Given a visible color image $I_{vi} \in \mathbb{R}^{H \times W \times 3}$ and a single-channel infrared image $I_{ir} \in \mathbb{R}^{H \times W \times 1}$, our goal is to produce a fused color image $I_{f} \in \mathbb{R}^{H \times W \times 3}$.

\paragraph{Preprocessing.}
We first convert the visible image from RGB to the YCbCr color space. This separates the luminance (Y channel) from the chrominance (Cb and Cr channels). The fusion is performed exclusively on the luminance channel $I_{vi}^Y$ and the infrared image $I_{ir}$, while the chrominance channels $I_{vi}^{CbCr}$ are preserved and directly passed to the final reconstruction stage, ensuring the color of the visible image is faithfully maintained.

\paragraph{U-Net for Guidance Map Generation.}
The core learnable component is a U-Net that generates a guidance map. We deliberately choose a classic convolutional U-Net~\cite{xu2020u2fusion} over more recent, complex backbones like Transformers or Restormers~\cite{Zamir2021Restormer, ma2022swinfusion}. While powerful, these architectures often incur prohibitive memory costs during full-resolution training, leading to the very patch-based compromises we aim to avoid. As shown in our ablation studies, even efficient full-convolution networks can lead to Out-Of-Memory (OOM) errors at large batch sizes. The U-Net's encoder-decoder structure with downsampling and skip connections is highly memory-efficient for dense prediction tasks. We design it with four downsampling stages, symmetrically mirroring the four decomposition levels of our Laplacian pyramid, creating an intuitive and synergistic link between the guidance generator and the fusion kernel. It takes the concatenated $I_{vi}^Y$ and $I_{ir}$ as input and produces a dense weight map $\mu \in [0, 1]^{H \times W}$, which serves as a per-pixel guidance signal for the subsequent fusion step.

\paragraph{Laplacian Pyramid Fusion Kernel.}
The actual fusion is performed by a fixed, non-learnable Laplacian pyramid fusion kernel. This choice is deliberate: the pyramid decomposition is a mathematically well-defined, interpretable process that naturally separates an image into multi-scale frequency bands. This structure is perfectly suited for guidance from a neural network, as a learned weight map can be seamlessly downsampled and applied at each pyramid level. The corresponding feature maps are combined using the guidance map $\mu$ as a linear weight:

\begin{equation}
    L_{fused}^k = (1 - \mu^k) \cdot L_{vi}^k + \mu^k \cdot L_{ir}^k
    \label{eq:fusion_rule}
\end{equation}
where $L^k$ denotes the $k$-th level of the Laplacian pyramid and $\mu^k$ is the resized guidance map. The final fused luminance channel $I_{fused}^Y$ is then reconstructed by collapsing the fused pyramid.

\paragraph{Reconstruction.}
Finally, the fused luminance $I_{fused}^Y$ is combined with the original chrominance channels $I_{vi}^{CbCr}$ and converted back from YCbCr to the RGB color space to produce the final output image $I_f$.

\subsection{Unsupervised Loss Function}
The U-Net is trained end-to-end using a comprehensive unsupervised loss function that requires no ground-truth fused images. This loss is composed of multiple terms designed to preserve crucial information from both the visible ($I_{vi}$) and infrared ($I_{ir}$) sources. The total loss $\mathcal{L}_{total}$ is a weighted sum of these components:
\begin{equation}
\begin{split}
    \mathcal{L}_{total} = \lambda_{max}\mathcal{L}_{max} + \lambda_{grad}\mathcal{L}_{grad} \\
    + \lambda_{ssim}\mathcal{L}_{ssim} + \lambda_{consist}\mathcal{L}_{consist}
\end{split}
\label{eq:total_loss}
\end{equation}
\begin{table*}[t]
    \caption{Quantitative comparison on MSRS, M3FD, and RoadScene. Total training time for our method is shown in parentheses. Data is sourced from original papers; efficiency details refer to Table \ref{tab:efficiency_analysis}.}
    \label{tab:main_comparison_ivf}
    \centering
    \resizebox{\textwidth}{!}{%
    \begin{tabular}{l|cccc|cccc|cccc|c}
        \toprule
        \multirow{2}{*}{Method} & \multicolumn{4}{c|}{MSRS} & \multicolumn{4}{c|}{M3FD} & \multicolumn{4}{c|}{RoadScene} & \multirow{2}{*}{\begin{tabular}[c]{@{}c@{}}Inference\\ Mode\end{tabular}} \\
        \cmidrule(lr){2-5} \cmidrule(lr){6-9} \cmidrule(lr){10-13}
        & EN$\uparrow$ & MI$\uparrow$ & VIF$\uparrow$ & $Q^{AB/F}$ $\uparrow$ & EN$\uparrow$ & MI$\uparrow$ & VIF$\uparrow$ & $Q^{AB/F}$ $\uparrow$ & EN$\uparrow$ & MI$\uparrow$ & VIF$\uparrow$ & $Q^{AB/F}$ $\uparrow$ & \\
        \midrule
        VisAsFused & 6.596 & 6.865 & 1.044 & 0.650 & 6.806 & 7.556 & 1.025 & 0.688 & 6.948 & 7.667 & 1.040 & 0.474 & - \\
        IrAsFused  & 5.309 & 5.668 & 1.028 & 0.384 & 7.130 & 8.070 & 1.019 & 0.367 & 7.563 & 8.469 & 1.061 & 0.665 & - \\
        \midrule
        SuperFusion \cite{TANG2022SuperFusion} & 6.587 & 3.596 & 0.813 & 0.557 & 6.726 & 4.345 & 0.664 & 0.522 & 6.990 & 3.562 & 0.608 & 0.452 & Full \\
        CDDFuse \cite{zhao2023cddfuse}     & 6.701 & 3.657 & 0.819 & 0.548 & \underline{7.070} & 3.994 & 0.802 & 0.613 & 7.120 & 3.001 & 0.610 & 0.450 & Full \\
        IFCNN \cite{zhang2020IFCNN}       & 5.975 & 1.706 & 0.579 & 0.479 & 6.935 & 2.630 & 0.685 & 0.590 & 7.222 & 2.842 & 0.591 & 0.536 & Full \\
        U2Fusion \cite{xu2020u2fusion}     & 5.246 & 2.183 & 0.512 & 0.391 & 6.872 & 2.683 & 0.673 & 0.578 & 6.739 & 2.578 & 0.564 & 0.506 & Full \\
        SwinFusion \cite{ma2022swinfusion}   & 6.619 & 3.652 & 0.825 & 0.558 & 6.844 & 4.020 & 0.746 & 0.616 & 7.000 & 3.334 & 0.614 & 0.450 & Full \\
        PSLPT \cite{wang2024pslpt}       & 6.307 & 2.284 & 0.753 & 0.553 & \textbf{7.204} & 4.563 & \textbf{0.958} & 0.321 & 7.077 & 2.001 & 0.134 & 0.171 & Full \\
        TC-MOA \cite{zhu2024tcmoa}      & 6.633 & 3.251 & 0.811 & 0.565 & 6.747 & 2.856 & 0.579 & 0.508 & 7.387 & 2.853 & 0.577 & 0.477 & Pad-640 \\
        DDFM \cite{zhao2023DDFM}        & \textbf{6.88}  & 2.35  & 0.81  & 0.58  & 6.86  & 2.52  & 0.81  & 0.49  & 7.41  & 2.35  & 0.75  & \textbf{0.65}  & Zero-shot \\
        Text-IF \cite{yi2024text}      & \underline{6.789} & \textbf{5.406} & 1.046 & 0.676 & 6.849 & \textbf{5.553} & 0.780 & 0.550 & 7.332 & \textbf{5.009} & 0.739 & 0.578 & Full+LLM \\
        DTPF \cite{dtpf}       & 6.749 & 4.883 & \underline{1.060} & \textbf{0.732} & 6.965 & 4.780 & 0.896 & \textbf{0.706} & 7.248 & 3.454 & 0.743 & 0.639 & Slide+LLM \\
        \midrule
        Ours (2e, 1.2min)   & 6.644 & 3.495 & 0.996 & 0.678 & 6.985 & 4.034 & 0.903 & 0.650 & 7.173 & 3.449 & \underline{0.830} & 0.513 & Full \\
        Ours (10e, 5.7min)  & 6.695 & 3.752 & 1.017 & 0.691 & 7.025 & 3.544 & 0.912 & 0.659 & \underline{7.439} & \underline{4.410} & \textbf{0.891} & \underline{0.649} & Full \\
        Ours (100e, 57min)  & 6.766 & \underline{4.890} & \textbf{1.079} & \underline{0.721} & 7.014 & \underline{5.177} & \underline{0.952} & \underline{0.696} & \textbf{7.458} & 3.845 & 0.822 & 0.646 & Full \\
        \bottomrule
    \end{tabular}%
    }
    
\end{table*}

Each component is detailed as follows:

\paragraph{Intensity Maximum Loss ($\mathcal{L}_{max}$)} encourages the fused image $I_f$ to retain the most significant intensity information from either source on a per-pixel basis. This is achieved by minimizing the L1 distance to the pixel-wise maximum of the source images' luminance channels ($I_{vi}^Y, I_{ir}$):
\begin{equation}
\mathcal{L}_{max} = \frac{1}{HW} \|I_f^Y - \max(I_{vi}^Y, I_{ir})\|_1
\end{equation}

\paragraph{Gradient Maximum Loss ($\mathcal{L}_{grad}$)} aims to preserve the most prominent edges and textural details. It penalizes the difference between the gradients of the fused image and the maximum gradients from the source images, where $\nabla$ is the Sobel gradient operator:
\begin{equation}
\mathcal{L}_{grad} = \frac{1}{HW} \|\nabla I_f^Y - \max(\nabla I_{vi}^Y, \nabla I_{ir})\|_1
\end{equation}

\paragraph{Structural Similarity Loss ($\mathcal{L}_{ssim}$)} ensures that the fused result maintains structural fidelity to both inputs. It is formulated as the sum of structural dissimilarity terms with respect to each source image:
\begin{equation}
\mathcal{L}_{ssim} = (1 - \text{SSIM}(I_f^Y, I_{vi}^Y)) + (1 - \text{SSIM}(I_f^Y, I_{ir}))
\end{equation}

\paragraph{Intensity Consistency Loss ($\mathcal{L}_{consist}$)} acts as a regularizer, ensuring the overall intensity distribution of the fused image does not excessively deviate from the source images. It is a sum of L1 losses to both inputs:
\begin{equation}
\mathcal{L}_{consist} = \frac{1}{HW} \left( \|I_f^Y - I_{vi}^Y\|_1 + \|I_f^Y - I_{ir}\|_1 \right)
\end{equation}

\begin{table*}[t]
    \caption{Comprehensive efficiency and parameter analysis. Previous methods use inconsistent training settings, making direct comparison unfair, but our method still performs best overall. Performance anchors on MSRS are provided; full metrics are in Table \ref{tab:main_comparison_ivf}.}
    \label{tab:efficiency_analysis}
    \centering
    \resizebox{\textwidth}{!}{%
    \begin{tabular}{lcccccccc}
        \toprule
        \multirow{2}{*}{Method} & \multicolumn{2}{c}{MSRS Performance} & \multirow{2}{*}{\begin{tabular}[c]{@{}c@{}}Training\\ Data\end{tabular}} & \multirow{2}{*}{\begin{tabular}[c]{@{}c@{}}Estimated Training\\ Time (Epochs)\end{tabular}} & \multirow{2}{*}{\begin{tabular}[c]{@{}c@{}}Inference\\ Mode\end{tabular}} & \multirow{2}{*}{\begin{tabular}[c]{@{}c@{}}MSRS Inf. Time\\ (361 images)\end{tabular}} & \multirow{2}{*}{\begin{tabular}[c]{@{}c@{}}Parameters\\ (M)\end{tabular}} \\
        \cmidrule(lr){2-3}
        & VIF$\uparrow$ & $Q^{AB/F}$ $\uparrow$ & & & & \\
        \midrule
        SuperFusion \cite{TANG2022SuperFusion} & 0.813 & 0.557 & MSRS Patch (256$\times$256) & $\sim$10h (1200 ep) & Full & 20min (bs. 4) & 11.23 \\
        CDDFuse \cite{zhao2023cddfuse}     & 0.819 & 0.548 & MSRS Patch (128$\times$128) & $\sim$1h (120 ep) & Full & $\sim$10min (bs. 1) & 2.40 \\
        IFCNN \cite{zhang2020IFCNN}       & 0.579 & 0.479 & MSRS Patch (128$\times$128) & $\sim$1h (120 ep) & Full & $\sim$40s (bs. 16) & 0.08 (+44.5) \\
        U2Fusion \cite{xu2020u2fusion}     & 0.512 & 0.391 & MSRS Patch (64$\times$64)   & $\sim$1h (120 ep) & Full & $\sim$1min (bs. 8) & 0.63 \\
        SwinFusion \cite{ma2022swinfusion}   & 0.825 & 0.558 & MSRS Patch (128$\times$128) & $\sim$2h (120 ep) & Full & $\sim$50min (bs. 4) & 13.04 \\
        PSLPT \cite{wang2024pslpt}       & 0.753 & 0.553 & MSRS Patch (128$\times$128 rep.) & $\sim$1h & Full & $\sim$30min (bs. 4) & 3.06 \\
        TC-MOA \cite{zhu2024tcmoa}      & 0.811 & 0.565 & MSRS Patch (448$\times$448) & $\sim$5h & Pad-640 & $\sim$30min (bs. 1) & 7.24 (+329) \\
        DDFM \cite{zhao2023DDFM}        & 0.81  & 0.58  & ILSVRC (Pre-trained) & Zero-shot & Full (100 step) & $\sim$1.6h (bs. 1) & (552.81) \\
        Text-IF \cite{yi2024text}      & 1.046 & 0.676 & 4 Datasets Patch (96$\times$96) & $>$10h & Full + LLM & $\sim$3h (bs. 1) & 63.8 (+151) \\
        DTPF \cite{dtpf}       & \underline{1.060} & \textbf{0.732} & MSRS Patch (128$\times$128) & 13.1h (500 ep) & Slide + LLM & 2.5h (bs. 1) & 40.3 (+151) \\
        \midrule
        Ours (2e)   & 0.996 & 0.678 & \textbf{MSRS Full Image} & \textbf{1.2min} (2 ep) & Full & \textbf{10s (bs. 16)} & 17.264 \\
        Ours (10e)  & 1.017 & 0.691 & \textbf{MSRS Full Image} & \underline{5.7min} (10 ep) & Full & \textbf{10s (bs. 16)} & 17.264 \\
        Ours (100e) & \textbf{1.079} & \underline{0.721} & \textbf{MSRS Full Image} & 57min (100 ep) & Full & \textbf{10s (bs. 16)} & 17.264 \\
        \bottomrule
    \end{tabular}%
    }
\end{table*}
The final performance is highly dependent on the balance between these loss terms, controlled by the weights $\lambda$. To find a robust and optimal configuration, we conducted an extensive grid search over 240 hyperparameter combinations. The final weights used for all experiments were selected from the high-performance region identified in this search, as detailed in our ablation studies and visualized in Figure~\ref{fig:loss_grid_search}.

\section{Experiments}
\label{sec:experiments}

We conduct various experiments to validate the efficiency, performance, and generalization capabilities of our hybrid fusion method. We first compare our approach against a wide range of state-of-the-art (SOTA) methods on several benchmark datasets, followed by detailed ablation studies to analyze key components and hardware advantages.

\begin{table*}[t]
  \centering
  \caption{Downstream detection performance on MSRS dataset. Unlike potentially deceptive fusion metrics that may favor source-copying, downstream tasks offer a more objective and stable performance benchmark.}
  \label{tab:detection}
  
  \resizebox{1.0\linewidth}{!}{
  \begin{tabular}{l|cccccccccc}
    \toprule
    \textbf{Metric / Method} & Vi (Base) & U2F \cite{xu2020u2fusion} & IFCNN \cite{zhang2020IFCNN} & SwinF \cite{ma2022swinfusion} & SuperF \cite{TANG2022SuperFusion} & CDD \cite{zhao2023cddfuse} & PSLPT \cite{wang2024pslpt} & Text-IF \cite{yi2024text} & DTPF \cite{dtpf} & \textbf{Ours} \\
    \midrule
    mAP@50    & 0.7955 & 0.9280 & 0.9210 & 0.9110 & 0.9210 & 0.9220 & 0.9150 & 0.9236 & \underline{0.9368} & \textbf{0.9518} \\
    mAP@50-95 & 0.5411 & 0.6570 & 0.6510 & 0.6410 & 0.6730 & 0.6620 & 0.5660 & 0.7052 & \underline{0.7058} & \textbf{0.7233} \\
    \bottomrule
  \end{tabular}}
  
\end{table*}

\subsection{Experimental Setup}

\paragraph{Datasets and Metrics.}

Our experiments span multiple fusion domains, using MSRS \cite{MSRS_dataset}, M3FD \cite{liu2022target}, and RoadScene \cite{roadscene} for infrared-visible fusion (IVF), alongside three medical tasks (PET, CT, SPECT) and the M3SVD \cite{TANG2025VideoFusion} video dataset. We employ a comprehensive suite of metrics: VIF \cite{VIF}, $Q^{AB/F}$ \cite{Qabf}, SSIM \cite{SSIM}, EN \cite{EN}, and MI \cite{MI}. Specifically, following standard multi-modal image fusion protocols, metrics like SSIM and VIF are calculated as the sum of preserved information from both source images: $SSIM_{total} = SSIM(I_{A},I_{f}) + SSIM(I_{B},I_{f})$. Consequently, these values range from 0 to 2, explaining why reported scores often exceed 1. While the main comparison table (Table \ref{tab:main_comparison_ivf}) reports the highest results from original papers to ensure fairness, our identical hardware comparisons rely on rigorous local reproduction.

\paragraph{Implementation Details.}
We use a U-Net with 4 downsampling stages as our guidance network. The model is trained using the AdamW optimizer with an initial learning rate of $1 \times 10^{-4}$. Critically, all training is performed on full-resolution images. As detailed in Table~\ref{tab:performance_benchmarks}, training is exceptionally efficient: one epoch on the MSRS dataset takes approximately 2.3 minutes on an RTX 4060 Laptop GPU and just 30 seconds on an RTX 4090. SOTA-comparable results are achieved within only 2 epochs. 

\subsection{Quantitative and Qualitative Comparisons}

\begin{figure*}[t]
  \centering
   \includegraphics[width=\textwidth]{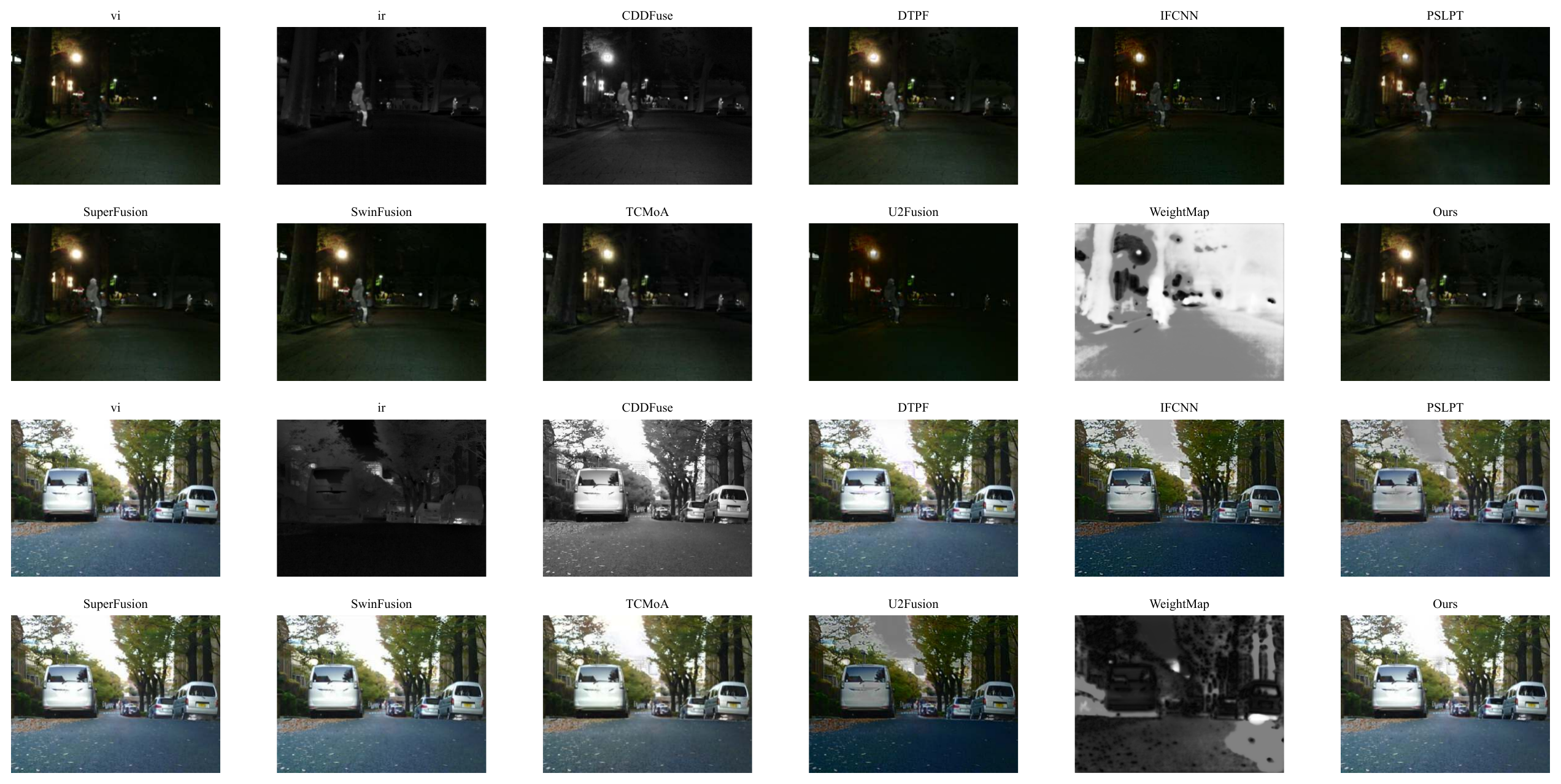}
   \caption{Qualitative comparison on the MSRS dataset. Our method effectively highlights the pedestrian from the infrared image while preserving the textural details from the visible image. The learned weight map validates this interpretable behavior. For more comparisons, please refer to supplementary materials. }
   \label{fig:ivf_comparison}
   
\end{figure*}

\paragraph{Infrared-Visible Fusion. }
Table~\ref{tab:main_comparison_ivf} presents a comprehensive comparison on three IVF datasets. The training configurations of different methods (taken from the original papers) vary significantly. Reporting only the highest performance metrics makes an unfair comparison against our method under their training conditions. Even so, our method, trained for mere minutes, achieves performance superior or comparable to SOTA models that require hours of patch-based training. For instance, after just 10 epochs (5.7 minutes), our model establishes new SOTA results on the RoadScene dataset. This table also reveals the practical burdens of other methods. Some, like DTPF~\cite{dtpf}, require a sliding window for inference due to high memory costs. Others, like TC-MOA~\cite{zhu2024tcmoa}, necessitate specific input padding (e.g., to $640 \times 512$), limiting flexibility. The reported inference batch sizes for competitors are often low (e.g., bs=1 for CDDFuse~\cite{zhao2023cddfuse}) due to VRAM constraints, whereas our method handles a large batch size of 16 with ease. Furthermore, some methods exhibit failure modes hidden by single metrics; PSLPT~\cite{wang2024pslpt}, for example, achieves a high VIF score on M3FD by essentially outputting the infrared image, as indicated by its collapsed Qabf score. Figure~\ref{fig:ivf_comparison} provides a visual comparison, where our method produces results with rich details and natural appearance. This strong performance extends to video fusion, as detailed in the supplementary materials. Our method is so efficient that we can directly apply the MSRS-trained model to the M3SVD test set without any video-specific training. This zero-shot performance is highly competitive with specialist methods like VideoFusion~\cite{TANG2025VideoFusion}, which are trained on video data. While our frame-by-frame approach may lack temporal consistency compared to dedicated video models, its strong zero-shot result further underscores the powerful generalization of our framework. 
\begin{table*}[t]
    \caption{Fair reproduction comparison on the MSRS dataset. All methods were re-trained on the same condition (RTX 4090, 4 batch size). Under the same training settings, our model undoubtedly achieves the best performance. `Time' denotes the training duration. Please refer to Table \ref{tab:inference_comparison}. for inference details. }
    \label{tab:reproduction_fair}
    \centering
    \resizebox{\textwidth}{!}{%
    \begin{tabular}{l|c|cccc|c|cccc}
        \toprule
        \multirow{2}{*}{Method} & \multicolumn{5}{c|}{1 Epoch (Rapid Convergence)} & \multicolumn{5}{c}{100 Epochs (Full Training)} \\
        \cmidrule(lr){2-6} \cmidrule(lr){7-11}
        & Time$\downarrow$ & EN$\uparrow$ & MI$\uparrow$ & VIF$\uparrow$ & $Q^{AB/F}$ $\uparrow$ & Time$\downarrow$ & EN$\uparrow$ & MI$\uparrow$ & VIF$\uparrow$ & $Q^{AB/F}$ $\uparrow$ \\
        \midrule
        U2Fusion \cite{xu2020u2fusion} & \textbf{28s} & 5.022 & 1.981 & 0.501 & 0.363 & \textbf{47min}  & 5.236 & 2.133 & 0.522 & 0.389 \\
        CDDFuse \cite{zhao2023cddfuse}  & 55s & 6.643 & 2.762 & 0.765 & 0.552 & 92min  & 6.711 & 2.897 & 0.811 & 0.623 \\
        Text-IF \cite{yi2024text}  & 57s & 6.541 & 2.782 & 0.746 & 0.523 & 97min  & 6.696 & 4.867 & 1.019 & 0.692 \\
        DTPF \cite{dtpf}     & 103s & 6.612 & 2.661 & 0.712 & 0.512 & 171min & 6.724 & 4.883 & 1.004 & 0.682 \\
        \midrule
        \textbf{Ours} & \underline{39s} & \textbf{6.771} & \textbf{3.499} & \textbf{0.996} & \textbf{0.678} & \underline{58min} & \textbf{6.766} & \textbf{4.890} & \textbf{1.079} & \textbf{0.721} \\
        \bottomrule
    \end{tabular}%
    }
    
\end{table*}

\paragraph{Downstream tasks. }
While our method achieves competitive scores across all metrics, we observe that traditional metrics can sometimes be deceptive; for instance, Entropy (EN) can be artificially high for noisy outputs, and VIF may favor models that simply copy one source image (Table\ref{tab:main_comparison_ivf}). We argue that the combination of VIF and $Q^{AB/F}$ provides a more stable reflection of fusion quality as they consistently track structural preservation and textural transfer and are reported in almost every related work. To provide an objective benchmark, we evaluate performance on a downstream object detection task using YOLOv8n. As shown in Table \ref{tab:detection}, our model consistently outperforms all SOTA competitors in mAP@50 and mAP@50-95, confirming that our fusion preserves critical semantic features more effectively.

\paragraph{Fair Reproduction on Identical Hardware.}
To address discrepancies in reporting, we reproduced four representative baselines on identical hardware (Table \ref{tab:reproduction_fair}). We select Text-IF \cite{yi2024text} and DTPF \cite{dtpf} as the strongest recent SOTA, while U2Fusion \cite{xu2020u2fusion} and CDDFuse \cite{zhao2023cddfuse} represent established efficient baselines. Notably, while CDDFuse has a relatively low parameter count, its Restormer-based architecture and dual-stage training significantly hinder throughput and consume excessive VRAM ($>48$G for $bs=16$)(Table \ref{tab:inference_comparison}). Under the same training settings, our model undoubtedly achieves the best performance. Our method achieves competitive metrics at only 1 epoch compared to SOTA models trained for 100 epochs, proving that the efficiency advantage of our model mainly lies in its rapid convergence, whereas it does not present a substantial speed gain under the same number of training epochs, with the efficiency primarily attributed to the CNN architecture being faster than the Restormer framework (Detailed in Table \ref{tab:inference_comparison}).

\begin{table*}[t]
    \caption{Resource comparison on an RTX 4090. Our method's VRAM usage scales gracefully with resolution. A batch size of 4 can be used in most consumer GPUs. }
    \centering
    \resizebox{\textwidth}{!}{
    \begin{tabular}{lccccccl}
        \toprule
        Method & Batch Size & Resolution & VRAM Usage(FP32 Inference) & Inference Time & Architecture \\
        \midrule
        CDDFuse & 16 & 640x480 & $>$48G & OOM & Restormer backbone \\
        CDDFuse & 1 & 640x480 & $\sim$10G & 399.1ms & Restormer backbone \\
        DTPF & 16 & 640x480 & $\sim$40G & 2483ms & XRestormer backbone \\
        U2 Fusion & 16 & 640x480 & $\sim$42G & 315.6ms & Full-size convolution \\
        \midrule
        \textbf{Ours} & \textbf{16} & \textbf{256x256} & \textbf{$\sim$0.8G ($\sim$2.7G BF16 Train)} & \textbf{266.66ms} & \textbf{UNet with downsampling convolution} \\
        \textbf{Ours} & \textbf{16} & \textbf{640x480} & \textbf{$\sim$12G ($\sim$19G BF16 Train)} & \textbf{429.93ms} & \textbf{UNet with downsampling convolution} \\
        \textbf{Ours} & \textbf{48} & \textbf{640x480} & \textbf{$\sim$43G (OOM BF16 Train)} & \textbf{862.16ms} & \textbf{UNet with downsampling convolution} \\
        \bottomrule
    \end{tabular}}
    \label{tab:inference_comparison}
\end{table*}

\begin{figure}[t]  
  \centering
  
   \includegraphics[width=\linewidth]{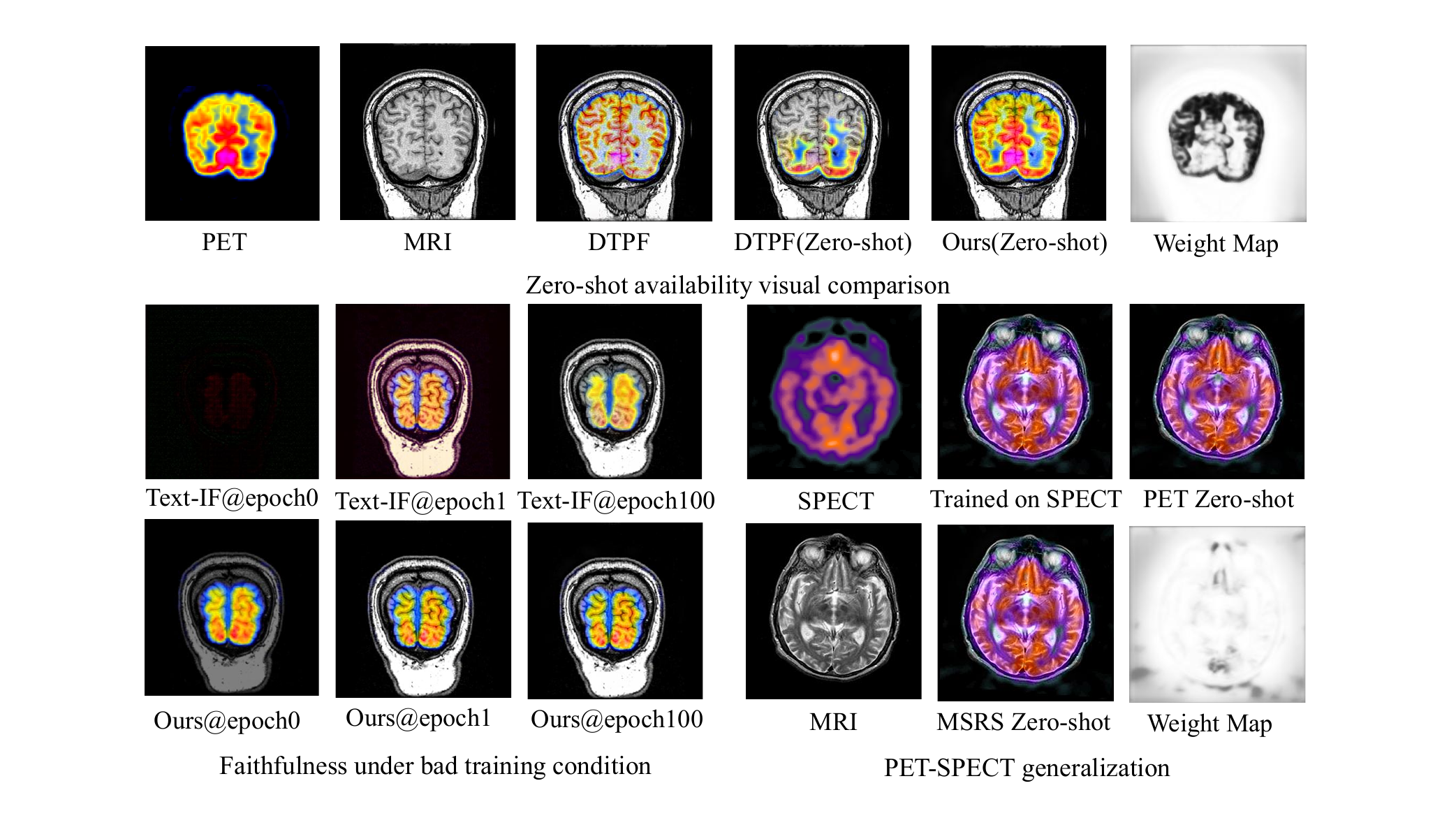}
   
   \caption{Qualitative zero-shot comparison on medical tasks. Our MSRS-trained model produces faithful fusion results. Competitor methods like DTPF can exhibit textural and color artifacts when generalizing, while our in-domain trained models (e.g., Trained on SPECT) also show strong cross-domain transferability (e.g., to PET).}
   \label{fig:zero_shot_medical}
   
\end{figure}

\begin{table*}[t]
    \caption{Quantitative comparison on medical image fusion datasets. }
    \label{tab:med_comparison}
    \centering
    \resizebox{\textwidth}{!}{%
    \setlength{\tabcolsep}{1.2pt}
    \begin{tabular}{l|ccccc|ccccc|ccccc|c}
        \toprule
        \multirow{2}{*}{Method} & \multicolumn{5}{c|}{PET-MRI} & \multicolumn{5}{c|}{CT-MRI} & \multicolumn{5}{c|}{SPECT-MRI} & \multirow{2}{*}{\begin{tabular}[c]{@{}c@{}}Training\\ Data\end{tabular}} \\
        \cmidrule(lr){2-6} \cmidrule(lr){7-11} \cmidrule(lr){12-16}
        & SSIM$\uparrow$ & EN$\uparrow$ & MI$\uparrow$ & VIF$\uparrow$ & $Q^{AB/F}$$\uparrow$ & SSIM$\uparrow$ & EN$\uparrow$ & MI$\uparrow$ & VIF$\uparrow$ & $Q^{AB/F}$$\uparrow$ & SSIM$\uparrow$ & EN$\uparrow$ & MI$\uparrow$ & VIF$\uparrow$ & $Q^{AB/F}$$\uparrow$ & \\
        \midrule
        MRI as Fused  & 1.251 & 1.927 & 3.090 & 1.017 & 0.215 & 1.390 & 2.559 & 3.356 & 1.027 & 0.543 & 1.288 & 3.436 & 3.709 & 1.026 & 0.238 & Input \\
        Med as Fused  & 1.251 & 4.302 & 4.831 & 1.046 & 0.815 & 1.390 & 3.220 & 4.017 & 1.025 & 0.528 & 1.288 & 3.515 & 4.229 & 1.043 & 0.784 & Input \\
        \midrule
        PSLPT \cite{wang2024pslpt} & 0.815 & 5.492 & 2.641 & 0.548 & 0.373 & 0.810 & 4.730 & 2.392 & 0.502 & 0.432 & 0.933 & 5.140 & 2.747 & 0.359 & 0.325 & Corr. \\
        EMFusion \cite{xu2021emfusion} & 1.221 & \underline{5.646} & 3.207 & 0.685 & 0.783 & 1.266 & 4.785 & 3.116 & 0.552 & 0.475 & 1.212 & 4.911 & 3.210 & 0.665 & 0.692 & Corr. \\
        MSRPAN \cite{MSRPAN} & 1.182 & 5.073 & 4.119 & 0.581 & \underline{0.799} & 1.261 & 4.202 & \textbf{4.126} & 0.436 & 0.455 & 1.153 & 4.753 & \underline{4.334} & 0.525 & 0.560 & Corr. \\
        SwinFusion \cite{ma2022swinfusion} & 0.725 & \textbf{5.817} & 3.662 & 0.703 & 0.683 & 0.579 & \underline{5.144} & 3.190 & 0.522 & 0.545 & 0.684 & \textbf{5.401} & 3.795 & 0.744 & 0.720 & Corr. \\
        Zero \cite{zero} & 1.162 & 5.495 & 3.786 & 0.635 & 0.774 & 1.199 & 4.406 & 3.358 & 0.320 & 0.582 & 1.180 & 4.997 & 3.564 & 0.582 & 0.681 & ImageNet \\
        U2Fusion \cite{xu2020u2fusion} & 0.494 & 5.532 & 2.785 & 0.460 & 0.292 & 0.042 & 4.896 & 1.694 & 0.074 & 0.489 & 0.479 & 4.539 & 2.870 & 0.419 & 0.696 & Corr. \\
        CDDFuse \cite{zhao2023cddfuse} & 1.227 & 5.149 & 3.572 & 0.650 & 0.765 & 1.224 & \textbf{5.733} & \underline{3.683} & 0.526 & 0.530 & 1.169 & 4.396 & 4.109 & 0.786 & 0.719 & Corr. \\
        Text-IF \cite{yi2024text} & \underline{1.232} & 5.443 & 3.718 & 0.640 & 0.690 & \underline{1.313} & 4.356 & 3.211 & 0.542 & 0.561 & 1.200 & 4.807 & 3.994 & 0.747 & 0.715 & Corr. \\
        DTPF \cite{dtpf} & 1.223 & 5.611 & \underline{4.248} & 0.909 & 0.782 & \underline{1.313} & 4.462 & 3.246 & \underline{0.641} & \underline{0.657} & \underline{1.210} & 5.035 & 4.248 & \underline{0.888} & \underline{0.746} & Corr. \\
        \midrule
        Ours (2e zero-shot) & 1.118 & 5.482 & 3.621 & \textbf{1.025} & 0.772 & 0.969 & 4.973 & 2.974 & \textbf{0.807} & 0.561 & \textbf{1.241} & \underline{4.608} & 3.568 & 0.840 & 0.732 & MSRS \\
        Ours (10e zero-shot) & 1.100 & 5.255 & \textbf{4.289} & 0.911 & 0.796 & 1.007 & 4.988 & 2.727 & \underline{0.701} & 0.622 & 1.214 & 4.603 & \textbf{4.482} & \textbf{0.965} & \underline{0.776} & MSRS \\
        \textbf{Ours (Trained)} & \textbf{1.253} & 4.985 & 4.016 & \underline{0.937} & \textbf{0.806} & \textbf{1.325} & 4.290 & 2.940 & 0.700 & \textbf{0.664} & \textbf{1.270} & 4.590 & 4.200 & \underline{0.956} & \textbf{0.780} & Corr. \\
        \bottomrule
    \end{tabular}%
    }
    
\end{table*}
\paragraph{Zero-shot on Medical Image Fusion.}
A key advantage of our method is its strong and faithful generalization, a critical requirement for medical imaging. As shown in Table~\ref{tab:med_comparison}, our model, trained \emph{only} on the MSRS natural scene dataset, achieves remarkable zero-shot performance on three medical fusion tasks. It even outperforms specialist models like EMFusion~\cite{xu2021emfusion} and the ImageNet-pretrained Zero~\cite{zero} method. Many SOTA methods like Text-IF and DTPF rely on large, frozen language models (+151M params) and were trained with extreme hardware (e.g., TC-MOA used 8xA6000 GPUs on large 448x448 patches). We argue that for real-world applications, it is more practical to address image degradation at the source (e.g., clean a camera lens) than to rely on such computationally expensive restoration solutions. Furthermore, as shown in Figure~\ref{fig:explain} and visualized in Figure~\ref{fig:zero_shot_medical}, reconstruction-based methods can introduce unfaithful artifacts. For example, DTPF's zero-shot result on PET-MRI exhibits textural changes and color shifts not present in the source images. Such ``hallucinations'' could lead to misdiagnosis. Our method avoids this by ensuring all fused information originates from the source images. This strong generalization stems from our model learning a fundamental allocation task, as evidenced in Table~\ref{tab:zeroshot_generalization}, where models trained on SPECT can zero-shot to PET and vice-versa, a capability that reconstruction-based models often lack. 

A core advantage of our hybrid design is its ``physical fallback'' mechanism, which is critical for clinical reliability. As visualized in Fig. \ref{fig:zero_shot_medical}, purely generative SOTA models often hallucinate or output noise under bad training conditions (Epoch 0,1) because they synthesize pixels from scratch. Conversely, our model acts as a traditional Laplacian fusion even with random weights, guaranteeing structural faithfulness and preventing misdiagnosis-prone artifacts. This robustness enables powerful zero-shot transfer, where a model trained only on MSRS natural scenes outperforms specialist medical models like EMFusion \cite{xu2021emfusion} and ImageNet-pretrained Zero-learning methods \cite{zero}. Our SPECT-PET cross-domain results (Table \ref{tab:zeroshot_generalization}) further confirm that the model learns a universal allocation policy rather than dataset-specific patterns. 

\begin{table}[h]
    \caption{Analysis of zero-shot generalization between medical tasks. Models demonstrate strong cross-domain performance, with the MSRS-trained model remaining highly competitive.}
    \centering
    \resizebox{\linewidth}{!}{
    \begin{tabular}{lcccccc}
        \toprule
        \multirow{2}{*}{Training Dataset (Knowledge Source)} & \multicolumn{3}{c}{Evaluation on \textbf{SPECT-MRI}} & \multicolumn{3}{c}{Evaluation on \textbf{PET-MRI}} \\
        \cmidrule(lr){2-4} \cmidrule(lr){5-7}
        & SSIM$\uparrow$ & VIF$\uparrow$ & Qabf$\uparrow$ & SSIM$\uparrow$ & VIF$\uparrow$ & Qabf$\uparrow$ \\
        \midrule
        \textbf{SPECT-MRI} (In-domain / Zero-shot) & \textbf{1.270} & \underline{0.956} & \textbf{0.780} & \textbf{1.259} & 0.898 & \underline{0.795} \\
        \textbf{PET-MRI} (Zero-shot / In-domain)   & \underline{1.233} & 0.861 & 0.754 & \underline{1.253} & \underline{0.937} & \textbf{0.806} \\
        MSRS (Zero-shot / Zero-shot)      & 1.214 & \textbf{0.965} & \underline{0.776} & 1.118 & \textbf{1.025} & 0.772 \\
        \bottomrule
    \end{tabular}}
    
    \label{tab:zeroshot_generalization}
\end{table}

\subsection{Ablation Studies and Analysis}

To dissect the effectiveness of our design choices, we conduct several ablation studies, analyzing the core paradigm, loss function, training dynamics, architectural choices, and the practical implications for hardware efficiency and reproducibility. For more please refer to supplementary materials. 

\paragraph{Effectiveness of the Hybrid Paradigm.}
We validate the hybrid design's core premise in Table~\ref{tab:comparison_hybrid}. A pure Laplacian fusion method is fast but yields low scores across datasets. In contrast, a pure deep learning model (our U-Net trained end-to-end for direct fusion) performs better but is far inferior to our hybrid approach, and even suffers from Out-Of-Memory (OOM) errors on full-resolution M3FD data—revealing pixel regression inefficiencies for this network. Our hybrid method achieves the highest scores by a large margin in only 2 epochs, confirming that synergy between the U-Net and classic fusion kernel is critical, enabling each component to excel at its specialized task.

\begin{table}[h]
    \caption{Ablation study comparing paradigms. The hybrid method significantly outperforms both alternatives.}
    \centering
    \resizebox{0.8\linewidth}{!}{
    \begin{tabular}{lcccccc}
        \toprule
        \multirow{2}{*}{Method} & \multicolumn{2}{c}{MSRS} & \multicolumn{2}{c}{M3FD} & \multicolumn{2}{c}{RoadScene} \\
        \cmidrule(lr){2-3} \cmidrule(lr){4-5} \cmidrule(lr){6-7}
        & VIF$\uparrow$ & Qabf$\uparrow$ & VIF$\uparrow$ & Qabf$\uparrow$ & VIF$\uparrow$ & Qabf$\uparrow$ \\
        \midrule
        \multicolumn{7}{l}{\textit{Traditional Methods}} \\
        Laplacian & 0.862 & 0.624 & 0.745 & 0.580 & 0.703 & 0.527 \\
        Heuristic weight map (W$\rightarrow$IR) & 0.887 & 0.622 & 0.748 & 0.573 & 0.714 & 0.519 \\
        \midrule
        \multicolumn{7}{l}{\textit{Pure Deep Learning (with ours unet)}} \\
        Ours (Directly Fuse, 200 epochs) & 0.981 & 0.362 & OOM & OOM & --- & --- \\
        Ours (Directly Fuse, 1000 epochs) & 0.868 & 0.570 & OOM & OOM & --- & --- \\
        \midrule
        \multicolumn{7}{l}{\textit{Our Hybrid Method}} \\
        Ours (2 epochs) & 0.997 & 0.674 & 0.827 & 0.605 & 0.787 & \textbf{0.581} \\
        Ours (10 epochs) & \textbf{1.032} & \textbf{0.694} & \textbf{0.924} & \textbf{0.664} & \textbf{0.808} & 0.556 \\
        \bottomrule
    \end{tabular}}
    
    \label{tab:comparison_hybrid}
\end{table}

\begin{figure}[h]
    \centering
    
    \includegraphics[width=\columnwidth]{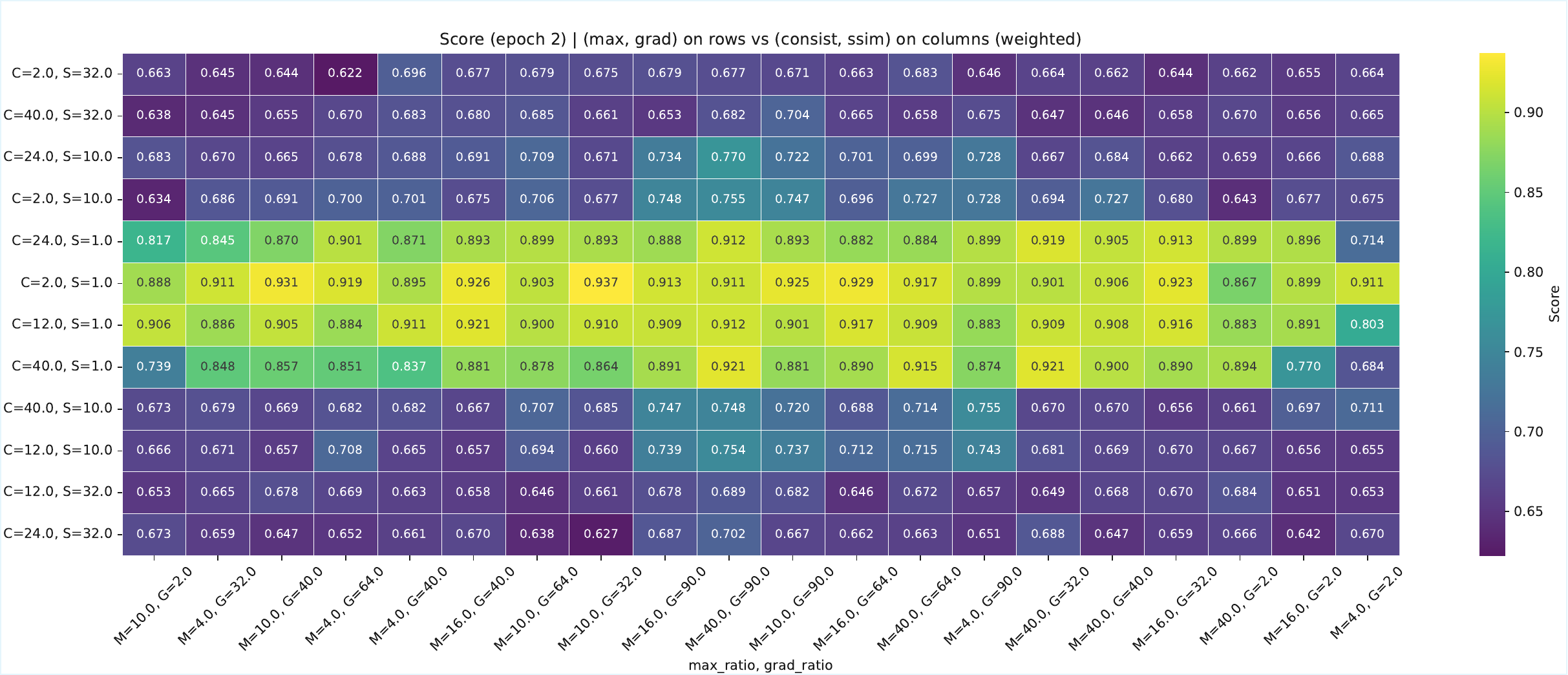}
    \caption{Hyperparameter grid search for the unsupervised loss function, showing a broad region of optimal performance.}
    \label{fig:loss_grid_search}
    
\end{figure}

\paragraph{Loss Function Hyperparameter Analysis.}
The unsupervised loss function is key to performance. We conducted a grid search over 240 hyperparameter combinations to find an optimal configuration. Figure~\ref{fig:loss_grid_search} visualizes MSRS results after 2 training epochs via a heatmap of composite reward score: $(\text{VIF} + 1.5 \times \text{Qabf} + \text{SSIM}) / 3$, balancing fusion quality assessment. The large contiguous bright yellow region indicates a broad, stable high-performance area, demonstrating our method is not overly sensitive to loss weights—ensuring robustness and easy deployment without exhaustive tuning. The final configuration for all experiments was selected from this optimal region.

\begin{figure}[h]
  \centering
  
   \includegraphics[width=\linewidth]{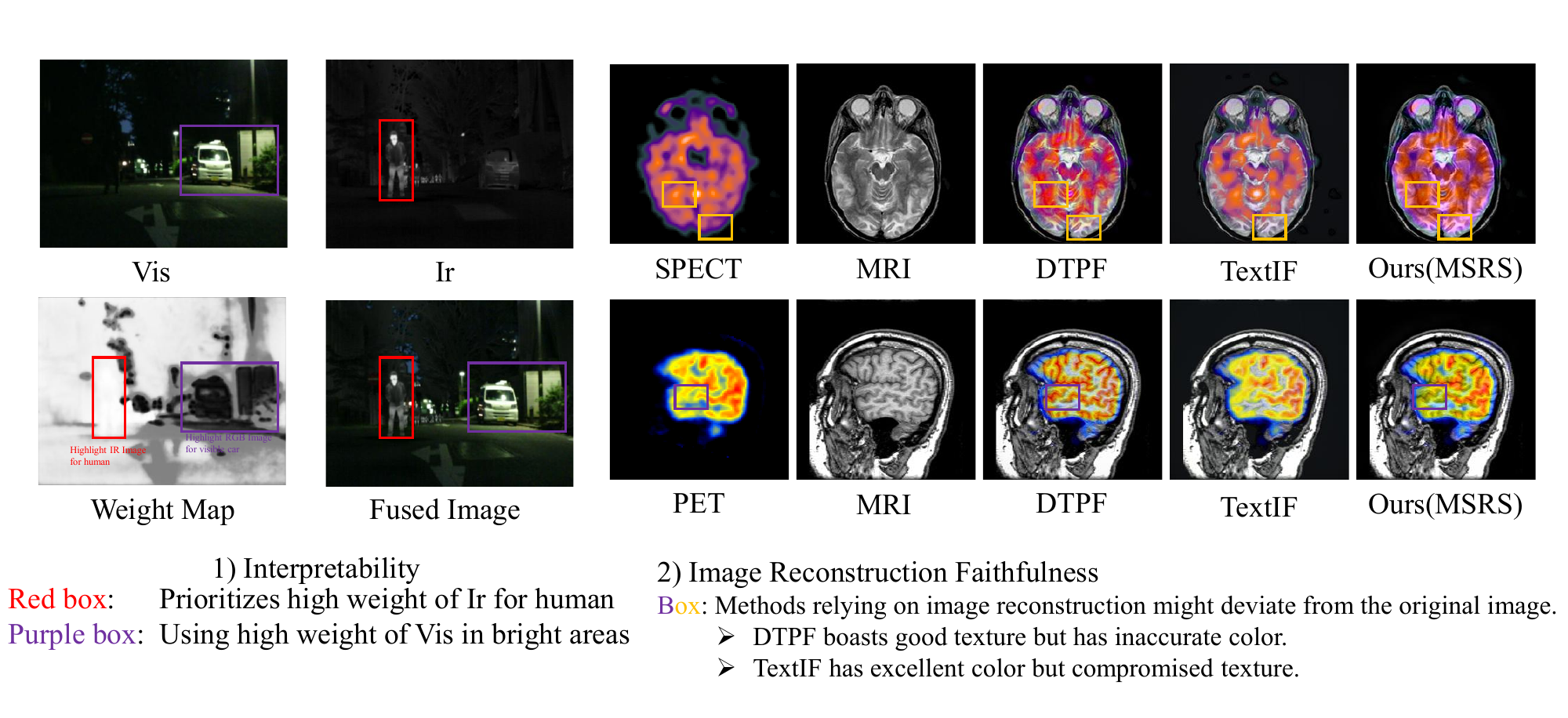}
   \caption{Interpretability of Hybrid method. For more please refer to supplementary materials. }
   
   \label{fig:explain}
\end{figure}

\paragraph{Interpretability and Faithfulness.}
Figure~\ref{fig:explain} showcases the interpretability and faithfulness of our method. The generated guidance map on the left provides a clear visualization of the model's allocation strategy: it assigns high weight (bright regions) to the infrared source to capture the salient human target, while assigning low weight (dark regions) to the background to preserve rich textural details from the visible source. This stands in stark contrast to reconstruction-based methods, which can introduce unfaithful artifacts. As conceptualized on the right, a model might alter the color of a lesion in a PET scan from yellow to red simply to enhance visual contrast, a ``hallucination'' that could lead to a severe misdiagnosis. Because our model's task is fundamentally one of allocation rather than generation, it is inherently more faithful to the source data, a crucial advantage for high-stakes applications.

\begin{table*}[t]
    \caption{Hardware benchmarks. SOTA results are rapidly attainable even on consumer-grade and free-tier hardware. *Performance on Windows was notably slower due to platform-specific overhead. }
    \centering
    \resizebox{\textwidth}{!}{%
    \begin{tabular}{l l cc cc ccr}
        \toprule
        \multirow{2}{*}{GPU} & \multirow{2}{*}{Environment} & BS & VRAM & Throughput & Time/Epoch & Epoch & \multicolumn{2}{c}{MSRS Metrics} \\
        \cmidrule(lr){8-9}
        & & & (est.) & (img/s) & (est.) & Used & VIF$\uparrow$ & Qabf$\uparrow$ \\
        \midrule
        RTX 4090 & Ubuntu & 16 & $\sim$19 GB & \textbf{22.1} & \textbf{$\sim$0.5 min} & 2 & 0.971 & 0.668 \\
        RTX 4060 Laptop & WSL2 Ubuntu & 4 & $\sim$5 GB & 7.8 & $\sim$2.3 min & 1 & 1.001 & 0.681 \\
        Google Colab T4 & Linux & 4 & $\sim$5 GB & 3.7 & $\sim$5.1 min & 1 & \textbf{1.016} & \textbf{0.691} \\
        RTX 4060 Laptop & Windows* & 4 & $\sim$5 GB & 1.0 & $\sim$18.8 min & 1 & 0.997 & 0.681 \\
        \bottomrule
    \end{tabular}%
    }
    
    \label{tab:performance_benchmarks}
\end{table*}

\begin{table}[t]
    \caption{Model scaling analysis. Larger models converge much faster with a minimal increase in inference time, making them more efficient.}
    \centering
    
    \resizebox{0.8\linewidth}{!}{
    \begin{tabular}{lccccc}
        \toprule
        Model Size & \begin{tabular}[c]{@{}c@{}}Inference Time \\ For bs. 16 $640 \times 480$ (ms)\end{tabular} & Epoch & MSRS VIF & MSRS Qabf & \begin{tabular}[c]{@{}c@{}}Total Training\\ Time (s)\end{tabular} \\
        \midrule
        \multirow{2}{*}{\textbf{17.264 M}} & \multirow{2}{*}{429.93} & 1 & 0.9028 & 0.6238 & 35 \\
        & & 2 & \textbf{1.0061} & \textbf{0.6798} & 69 \\
        \midrule
        \multirow{3}{*}{\textbf{608.642 K}} & \multirow{3}{*}{329.17} & 1 & 0.7022 & 0.4306 & 25 \\
        & & 2 & 0.7621 & 0.5583 & 48 \\
        & & 6 & 0.8094 & 0.5965 & 140 \\
        \midrule
        \multirow{1}{*}{\textbf{80.558 K}} & \multirow{1}{*}{272.23} & 1 & 0.6926 & 0.3930 & 25 \\
        \bottomrule
    \end{tabular}}
    
    \label{tab:model_convergence}
\end{table}

\paragraph{Hardware Efficiency and Model Scaling.}
A core motivation for our work is to democratize high-performance fusion. Table~\ref{tab:performance_benchmarks} provides concrete evidence, showing that SOTA-comparable results are attainable in minutes on accessible hardware like an RTX 4060 Laptop or a free-tier Google Colab T4 GPU. This efficiency is further explained in Table~\ref{tab:inference_comparison}. At full resolution, the primary VRAM bottleneck is the size of the feature maps, not the model's parameter count. By using a U-Net with downsampling, our memory usage remains manageable ($\sim$12GB for $640 \times 480$ inference), unlike competitors with Restormer or full-convolution backbones that require over 40GB of VRAM and face OOM errors. This is further supported by our model scaling analysis in Table~\ref{tab:model_convergence}. Increasing our model's parameter count by over 200x (from 80k to 17M) has only a minor impact on inference time. However, the larger model converges significantly faster, achieving superior metrics in fewer epochs. This reveals a key insight: for our architecture, a larger U-Net is paradoxically more training-efficient, as it learns the guidance task much more quickly.

\section{Conclusion}
\label{sec:conclusion}
We presented a hybrid framework that resolves the efficiency-performance trade-off in image fusion. Our method decouples a learnable U-Net guide from a fixed Laplacian pyramid kernel, which eliminates the train-inference gap and enables highly efficient, full-resolution training. This design achieves state-of-the-art comparable results in minutes on consumer hardware across diverse domains while ensuring strong zero-shot generalization and faithfulness to the source inputs. 


\clearpage

%
%
\bibliographystyle{unsrt}
\bibliography{main}

\begin{thebibliography}{10}

\bibitem{Albarqaan2024Image}
Hessah Albarqaan, Rongjun Qin, and Yang Teng.
\newblock {Image Fusion in Remote Sensing: An Overview and Meta-Analysis}.
\newblock {\em Photogrammetric Engineering \& Remote Sensing}, 90(12):755--775, December 2024.

\bibitem{MI}
Jiayi Ma, Yong Ma, and Chang Li.
\newblock Infrared and visible image fusion methods and applications: A survey.
\newblock {\em Information Fusion}, 45:153--178, 2019.

\bibitem{xu2021emfusion}
Han Xu and Jiayi Ma.
\newblock Emfusion: An unsupervised enhanced medical image fusion network.
\newblock {\em Information Fusion}, 2021.

\bibitem{zhao2023cddfuse}
Zixiang Zhao, Haowen Bai, Jiangshe Zhang, Yulun Zhang, Shuang Xu, Zudi Lin, Radu Timofte, and Luc Van~Gool.
\newblock Cddfuse: Correlation-driven dual-branch feature decomposition for multi-modality image fusion.
\newblock In {\em Proceedings of the IEEE/CVF conference on computer vision and pattern recognition}, pages 5906--5916, 2023.

\bibitem{laplacianPyramidFusionNetwork}
Jiaxin Yao, Yongqiang Zhao, Yuanyang Bu, Seong~G. Kong, and Jonathan Cheung-Wai Chan.
\newblock Laplacian pyramid fusion network with hierarchical guidance for infrared and visible image fusion.
\newblock {\em IEEE Transactions on Circuits and Systems for Video Technology}, 33(9):4630--4644, 2023.

\bibitem{TANG2022SuperFusion}
Linfeng Tang, Yuxin Deng, Yong Ma, Jun Huang, and Jiayi Ma.
\newblock Superfusion: A versatile image registration and fusion network with semantic awareness.
\newblock {\em IEEE/CAA Journal of Automatica Sinica}, 9(12):2121--2137, 2022.

\bibitem{zhang2020IFCNN}
Yu~Zhang, Yu~Liu, Peng Sun, Han Yan, Xiaolin Zhao, and Li~Zhang.
\newblock Ifcnn: A general image fusion framework based on convolutional neural network.
\newblock {\em Information Fusion}, 54:99--118, 2020.

\bibitem{ma2019fusionGAN}
Jiayi Ma, Wei Yu, Pengwei Liang, Chang Li, and Junjun Jiang.
\newblock Fusiongan: A generative adversarial network for infrared and visible image fusion.
\newblock {\em Information Fusion}, 48:11--26, 2019.

\bibitem{xu2020u2fusion}
Han Xu, Jiayi Ma, Junjun Jiang, Xiaojie Guo, and Haibin Ling.
\newblock U2fusion: A unified unsupervised image fusion network.
\newblock {\em IEEE Transactions on Pattern Analysis and Machine Intelligence}, 44(1):502--518, 2020.

\bibitem{zhao2023DDFM}
Zixiang Zhao, Haowen Bai, Yuanzhi Zhu, Jiangshe Zhang, Shuang Xu, Yulun Zhang, Kai Zhang, Deyu Meng, Radu Timofte, and Luc Van~Gool.
\newblock Ddfm: Denoising diffusion model for multi-modality image fusion.
\newblock In {\em Proceedings of the IEEE/CVF International Conference on Computer Vision (ICCV)}, pages 8082--8093, October 2023.

\bibitem{ma2022swinfusion}
Jiayi Ma, Linfeng Tang, Fan Fan, Jun Huang, Xiaoguang Mei, and Yong Ma.
\newblock Swinfusion: Cross-domain long-range learning for general image fusion via swin transformer.
\newblock {\em IEEE/CAA Journal of Automatica Sinica}, 9(7):1200--1217, 2022.

\bibitem{yi2024text}
Xunpeng Yi, Han Xu, Hao Zhang, Linfeng Tang, and Jiayi Ma.
\newblock Text-if: Leveraging semantic text guidance for degradation-aware and interactive image fusion.
\newblock In {\em Proceedings of the IEEE/CVF Conference on Computer Vision and Pattern Recognition (CVPR)}, 2024.

\bibitem{dtpf}
Ran Zhang, Xuanhua He, Ke~Cao, Liu Liu, Li~Zhang, Man Zhou, and Jie Zhang.
\newblock Distilling textual priors from llm to efficient image fusion, 2025.

\bibitem{laplacianCNN}
Jun Fu, Weisheng Li, Jiao Du, and Bin Xiao.
\newblock Multimodal medical image fusion via laplacian pyramid and convolutional neural network reconstruction with local gradient energy strategy.
\newblock {\em Computers in Biology and Medicine}, 126:104048, 11 2020.

\bibitem{Zamir2021Restormer}
Syed~Waqas Zamir, Aditya Arora, Salman Khan, Munawar Hayat, Fahad~Shahbaz Khan, and Ming-Hsuan Yang.
\newblock Restormer: Efficient transformer for high-resolution image restoration.
\newblock In {\em CVPR}, 2022.

\bibitem{wang2024pslpt}
Wu~Wang, Liang-Jian Deng, and Gemine Vivone.
\newblock A general image fusion framework using multi-task semi-supervised learning.
\newblock {\em Information Fusion}, page 102414, 2024.

\bibitem{zhu2024tcmoa}
Pengfei Zhu, Yang Sun, Bing Cao, and Qinghua Hu.
\newblock Task-customized mixture of adapters for general image fusion.
\newblock In {\em Proceedings of the IEEE/CVF Conference on Computer Vision and Pattern Recognition}, 2024.

\bibitem{MSRS_dataset}
Linfeng Tang, Jiteng Yuan, Hao Zhang, Xingyu Jiang, and Jiayi Ma.
\newblock {MSRS: Multi-Spectral Road Scenarios for Practical Infrared and Visible Image Fusion}.
\newblock \url{https://github.com/Linfeng-Tang/MSRS}, 2022.

\bibitem{liu2022target}
Jinyuan Liu, Xin Fan, Zhanbo Huang, Guanyao Wu, Risheng Liu, Wei Zhong, and Zhongxuan Luo.
\newblock Target-aware dual adversarial learning and a multi-scenario multi-modality benchmark to fuse infrared and visible for object detection.
\newblock In {\em Proceedings of the IEEE/CVF Conference on Computer Vision and Pattern Recognition}, pages 5802--5811, 2022.

\bibitem{roadscene}
Han Xu, Jiayi Ma, Zhuliang Le, Junjun Jiang, and Xiaojie Guo.
\newblock Fusiondn: A unified densely connected network for image fusion.
\newblock In {\em proceedings of the Thirty-Fourth AAAI Conference on Artificial Intelligence}, 2020.

\bibitem{TANG2025VideoFusion}
Linfeng Tang, Yeda Wang, Meiqi Gong, Zizhuo Li, Yuxin Deng, Xunpeng Yi, Chunyu Li, Han Xu, Hao Zhang, and Jiayi Ma.
\newblock Videofusion: A spatio-temporal collaborative network for multi-modal video fusion and restoration.
\newblock {\em arXiv preprint arXiv:2503.23359}, 2025.

\bibitem{VIF}
Yu~Han, Yunze Cai, Yin Cao, and Xiaoming Xu.
\newblock A new image fusion performance metric based on visual information fidelity.
\newblock {\em Information Fusion}, 14(2):127--135, 2013.

\bibitem{Qabf}
G.~Piella and H.~Heijmans.
\newblock A new quality metric for image fusion.
\newblock In {\em Proceedings 2003 International Conference on Image Processing (Cat. No.03CH37429)}, volume~3, pages III--173, 2003.

\bibitem{SSIM}
Zhou Wang, A.C. Bovik, H.R. Sheikh, and E.P. Simoncelli.
\newblock Image quality assessment: from error visibility to structural similarity.
\newblock {\em IEEE Transactions on Image Processing}, 13(4):600--612, 2004.

\bibitem{EN}
J.~Wesley Roberts, Jan~A. van Aardt, and Fethi~Babikker Ahmed.
\newblock {Assessment of image fusion procedures using entropy, image quality, and multispectral classification}.
\newblock {\em Journal of Applied Remote Sensing}, 2(1):023522, 2008.

\bibitem{MSRPAN}
Jun Fu, Weisheng Li, Jiao Du, and Yuping Huang.
\newblock A multiscale residual pyramid attention network for medical image fusion.
\newblock {\em Biomedical Signal Processing and Control}, 66:102488, 2021.

\bibitem{zero}
Fayez Lahoud and Sabine Süsstrunk.
\newblock Zero-learning fast medical image fusion.
\newblock In {\em 2019 22th International Conference on Information Fusion (FUSION)}, pages 1--8, 2019.

\end{thebibliography}
\end{document}